\documentclass[11pt]{article}

\usepackage{acl}

\usepackage{graphicx}
\usepackage{multirow}
\usepackage{url}

\begin{document}
\title{Knowledge Graph Representations for LLM-Based Policy Compliance Reasoning}

\author{
  Wilder Baldwin \\
  University of Maine \\
  \texttt{wilder.baldwin@maine.edu} \\
  \And
  Sepideh Ghanavati \\
  University of Maine \\
  \texttt{sepideh.ghanavati@maine.edu} \\
}

\date{February 2026}

\maketitle

\begin{abstract}
The risks posed by AI features are increasing as they are rapidly integrated into software applications. In response, regulations and standards for safe and secure AI have been proposed. In this paper, we present an agentic framework that constructs knowledge graphs (KGs) from AI policy documents and retrieves policy-relevant information to answer questions. We build KGs from three AI risk-related polices under two ontology schemas, and then evaluate five LLMs on 42 policy QA tasks spanning six reasoning types, from entity lookup to cross-policy inference, using both heuristic scoring and an LLM-as-judge. KG augmentation improves scores for all five models, and an open, LLM-discovered schema matches or exceeds the formal ontology. 
\end{abstract}

\section{Introduction}
\label{sec:introduction}

% Identifying policy, security, and privacy related risks in AI systems is a new and developing problem in the face of new AI governance frameworks and regulations such as the EU AI Act \citep{euaiact2024}, and security standards such as NIST AI RMF \citep{tabassi2023nist}, OWASP Top 10 for LLM Applications \citep{owasp2024llmtop10}, amongst others, which creates a need for automated tools that help practitioners navigate overlapping and evolving requirements.
AI governance frameworks such as the EU AI Act~\citep{euaiact2024}, the NIST AI RMF~\citep{tabassi2023nist}, and the OWASP Top 10 for LLM Applications~\citep{owasp2024llmtop10} impose overlapping requirements on the development and deployment of AI systems. Practitioners must navigate these evolving and often cross-referencing obligations, creating a need for automated tools that can identify policy, security, and privacy risks in AI systems.

Large language models (LLMs) are increasingly applied to related tasks, from extracting obligations from legislative text \citep{galli2025aiact} to verifying code-level compliance \citep{chung2025graphcompliance}. Knowledge graphs (KGs) pair well with LLMs to decompose dense policy texts into typed entities and relations that preserve the structure and provenance of policy requirements.

Recent work shows that KG-augmented LLM systems outperform text-only baselines on compliance tasks \citep{chung2025graphcompliance}, and the method of structuring the KG and presenting it to an LLM---from ontology design to serialization format--- affects reasoning quality \citep{markowitz2025kgllmbench}. However, these findings were established on general-purpose KGs (Wikidata, Freebase) and open-domain QA.
% that serialization format can cause up to 17.5\% absolute performance differences on KG comprehension benchmarks \citep{markowitz2025kgllmbench}.

In this paper, we propose an agentic AI framework that includes KGs from three AI governance documents using an end-to-end pipeline: an LLM chunking agent segments policy text via a scan-and-review process, a schema-constrained extractor produces typed entities and relations, then an agentic retrieval engine traverses the resulting graph to curate evidence for answer synthesis.
We evaluate different ontology schemas---a closed standards-derived schema grounded in the AI Risk Ontology (AIRO) \citep{golpayegani2022airo}, and an open schema with emergent type labels---across three retrieval conditions (no context, full graph serialization, and agent-curated traversal) on 42 policy QA tasks spanning six reasoning types. Our evaluation is grounded in both KG comprehension benchmarks \citep{markowitz2025kgllmbench} and legal NLP task taxonomies \citep{guha2023legalbench}, ranging from single-hop entity lookup (T1) to cross-policy reasoning (T6).
Evaluation uses both heuristic metrics matched to each task type and a multi-criteria LLM-as-Judge pipeline \cite{kim2024prometheus, liu2023geval}.
We center our findings around two research questions: 

-\textbf{RQ1} Can KGs integrate with LLMs to improve their ability to answer policy-related questions about risk in AI systems?  
We find that KG augmentation improves judge scores across all five models (+0.17 to +0.55), with the largest gains on tasks requiring verbatim policy citations, which LLMs cannot reliably produce from parametric knowledge alone.

-\textbf{RQ2:} How can we optimize the KG creation and retrieval process for agentic Q\&A in this domain?
We find that an open, LLM-discovered ontology matches or exceeds a formal standards-derived schema,\ but agentic graph traversal degrades performance for smaller models.
\section{Related Work}

\subsection{LLM-Based Knowledge Graph Construction and Retrieval}

Recent work demonstrates that LLMs can automate KG construction through multi-stage pipelines (i.e., extracting open triples, defining schemas, and canonicalizing entities) \cite{zhang2024edc, mo2025kggen, anuyah2025codekg}, with CoDe-KG achieving 92.4\% F1 on gold-annotated triples \cite{anuyah2025codekg}. \cite{zhu2024kgconstruction} evaluate LLMs across eight KG tasks and introduce AutoKG, a multi-agent approach that coordinates extraction and reasoning stages, finding that LLMs are better suited as inference assistants than zero-shot extractors.

% To address KG retrieval, \cite{edge2024graphrag} propose Graph RAG, which builds entity knowledge graphs from source documents, pregenerates community summaries for related entity groups, and consolidates partial responses at query time, yielding improvements over conventional RAG for queries over large corpora. \cite{luo2024rog} develop RoG, a planning-retrieval-reasoning framework where the LLM generates relation paths grounded in KG structure, retrieves valid reasoning paths, and then conducts inference using those paths. \citet{li2025subgraphrag} propose SubgraphRAG, which retrieves lightweight subgraphs using a parallel triple-scoring mechanism and feeds them to LLMs without fine-tuning.  \cite{zhu2024kgconstruction} evaluate LLMs across eight datasets, tasked with entity, relation, and event extraction, as well as linking and inferencing KGs for reasoning. They introduce AutoKG, a multi-agent approach for KG construction that coordinates extraction and reasoning stages. \citet{ma2025kgqasurvey} provide a comprehensive survey of the KG+LLM integration landscape for question answering, proposing a taxonomy that categorizes methods by QA type and KG role, and concluding that KG integration mitigates core LLM limitations including poor multi-hop reasoning, outdated knowledge, and hallucination. 

For LLM-augmented  KG-retrieval, Graph RAG \cite{edge2024graphrag} pregenerates community summaries over entity graphs for query-time consolidation. RoG \cite{luo2024rog} generates relation paths grounded in KG structure for multi-hop inference, while SubgraphRAG \cite{li2025subgraphrag} retrieves lightweight subgraphs via parallel triple scoring without fine-tuning.
\citet{ma2025kgqasurvey} survey the KG+LLM landscape and conclude that KG integration mitigates core LLM limitations, including poor multi-hop reasoning, outdated knowledge, and hallucination.

An open question is how to represent graph-structured knowledge as text. \citet{markowitz2025kgllmbench} evaluate five serialization strategies on KG comprehension tasks and find that structured JSON is optimal. \citet{dai2024kgunderstand} find that linearized triples outperform natural language descriptions for fact-intensive KG questions, and \citet{wu2024thinkingkg} show that code-based KG representations improve complex reasoning over both natural language and triple formats.

\subsection{Regulatory Compliance and NLP}

% A growing body of work applies LLMs to regulatory compliance and policy question answering. \citet{chung2025graphcompliance} propose GraphCompliance, which decomposes regulatory text into a Policy Graph---encoding compliance units as subject-constraint-context-condition tuples linked via cross-reference edges---and aligns it with a Context Graph representing runtime scenarios as SPO triples. \citet{agarwal2025ragulating} propose RAGulating Compliance, an agentic system that extracts ontology-free SPO triplets from regulatory documents, embeds them alongside text and metadata in a unified vector database, and uses orchestrated agents with triplet-level retrieval for QA. \citet{kovari2025chataiact} develop an AI-driven chatbot using RAG for EU AI Act compliance verification, comparing naive versus graph-based RAG approaches and integrating both public and proprietary standards. \citet{garza2024privcompkg} propose PrivComp-KG, combining RAG with a GDPR knowledge graph and SWRL rule checks for compliance verification.

A growing body of work applies structured representations to regulatory compliance QA. GraphCompliance \cite{chung2025graphcompliance} decomposes regulatory text into policy graphs encoding compliance units as subject-constraint-context-condition tuples. RAGulating Compliance \cite{agarwal2025ragulating} extracts ontology-free SPO triplets and uses orchestrated agents with triplet-level retrieval. \citet{kovari2025chataiact} compare naive versus graph-based RAG for EU AI Act compliance, and PrivComp-KG \cite{garza2024privcompkg} combines a GDPR KG with SWRL rule checks. These frameworks validate graph-based approaches to compliance but evaluate them on single policies; our work extends to cross-policy reasoning across three frameworks.

% Our evaluation draws on established benchmarks from both KG question answering and legal NLP. In the KG domain, \citet{markowitz2025kgllmbench} define five comprehension tasks---triple retrieval, shortest path, aggregation by relation, neighbor property aggregation, and highest degree node. For multi-hop reasoning, HotpotQA \citep{yang2018hotpotqa} established a standard for questions requiring reasoning across multiple documents with explicit supporting evidence, while MuSiQue \citep{trivedi2022musique} introduced compositional multi-hop questions designed to resist shortcut reasoning, requiring genuine multi-step inference. In legal and policy related NLP, LegalBench \citep{guha2024legalbench} provides 162 tasks organized into six categories of legal reasoning: issue spotting, rule recall, rule application, rule conclusion, interpretation, and rhetorical understanding.

In the legal and policy domain, \citet{galli2025aiact} apply LLaMA 3.3 70B with traditional NLP in a four-stage pipeline to extract obligations from the EU AI Act, constructing searchable KGs. \citet{colombo2024legislative} develop a platform combining legislative KGs with LLMs for Italian legislation, and observe that KGs provide the structured legal information LLMs need to avoid hallucinations. \citet{cui2023poligraph} extract KGs from privacy policies, identify 40\% more data collection statements than prior methods, and enable contradiction detection.

% \subsection{Policy Compliance and NLP}

% \paragraph{Corpora and datasets.} \cite{wilson2016opp115} introduce OPP-115, a corpus of 115 privacy policies with 23K fine-grained data practice annotations by law students. This dataset remains one of the most popular for textual classification of privacy policies. \cite{ahmad2020policyqa} build on this with PolicyQA, providing 25,017 reading comprehension examples with 714 human-annotated questions framed as span extraction. \cite{ravichander2019privacyqa} take a complementary approach with PrivacyQA, collecting 1,750 crowdworker questions about mobile app privacy policies paired with expert-annotated answers, finding that strong neural networks based solutions for classification still underperformed humans by ${\sim}$0.3 F1. \cite{ahmad2021policyie} extend structured extraction with PolicyIE, annotating 5,250 intents and 11,788 slots across 31 policies and showing that Seq2Seq outperforms sequence tagging for slot filling. \cite{srinath2021privaseer} release PrivaSeer, a corpus of over one million English privacy policies collected via web crawling, which they use to pretrain PrivBERT, a domain-specific BERT that achieves state-of-the-art on data practice classification and question answering. \cite{marottawurgler2025longtext} construct a long-text privacy policy corpus annotated across 64 dimensions reflecting both EU and US privacy regulation, capturing challenges specific to legal language such as indeterminacy, clause interdependence, and meaningful silence.

Several corpora and benchmarks support policy NLP research. OPP-115 \cite{wilson2016opp115} provides 23K annotated privacy policy data practices, extended by PolicyQA \cite{ahmad2020policyqa} for reading comprehension and PolicyIE \cite{ahmad2021policyie} for structured extraction. Domain-specific pretraining consistently improves performance: PrivaSeer \cite{srinath2021privaseer} and PrivacyGLUE \cite{shankar2023privacyglue} confirm that privacy policies constitute a distinct language domain requiring specialized models. LegalBench \cite{guha2023legalbench} provides 162 tasks covering six types of legal reasoning. Our taxonomy draws on both KG comprehension benchmarks \cite{markowitz2025kgllmbench} and extends these legal NLP taxonomies, with cross-policy reasoning tasks (T6) not previously addressed.
\section{Methodology}
\label{sec:methodology}

 %Our tool-support framework comprises a four-stage pipeline: (1)~document chunking, (2)~ontology-constrained knowledge graph extraction, (3)~adaptive agentic retrieval over the resulting graph, and (4)~answer synthesis with grounded evaluation.

\begin{figure*}[t]
\centering
\includegraphics[width=0.65\textwidth,height=0.63\textheight,keepaspectratio]{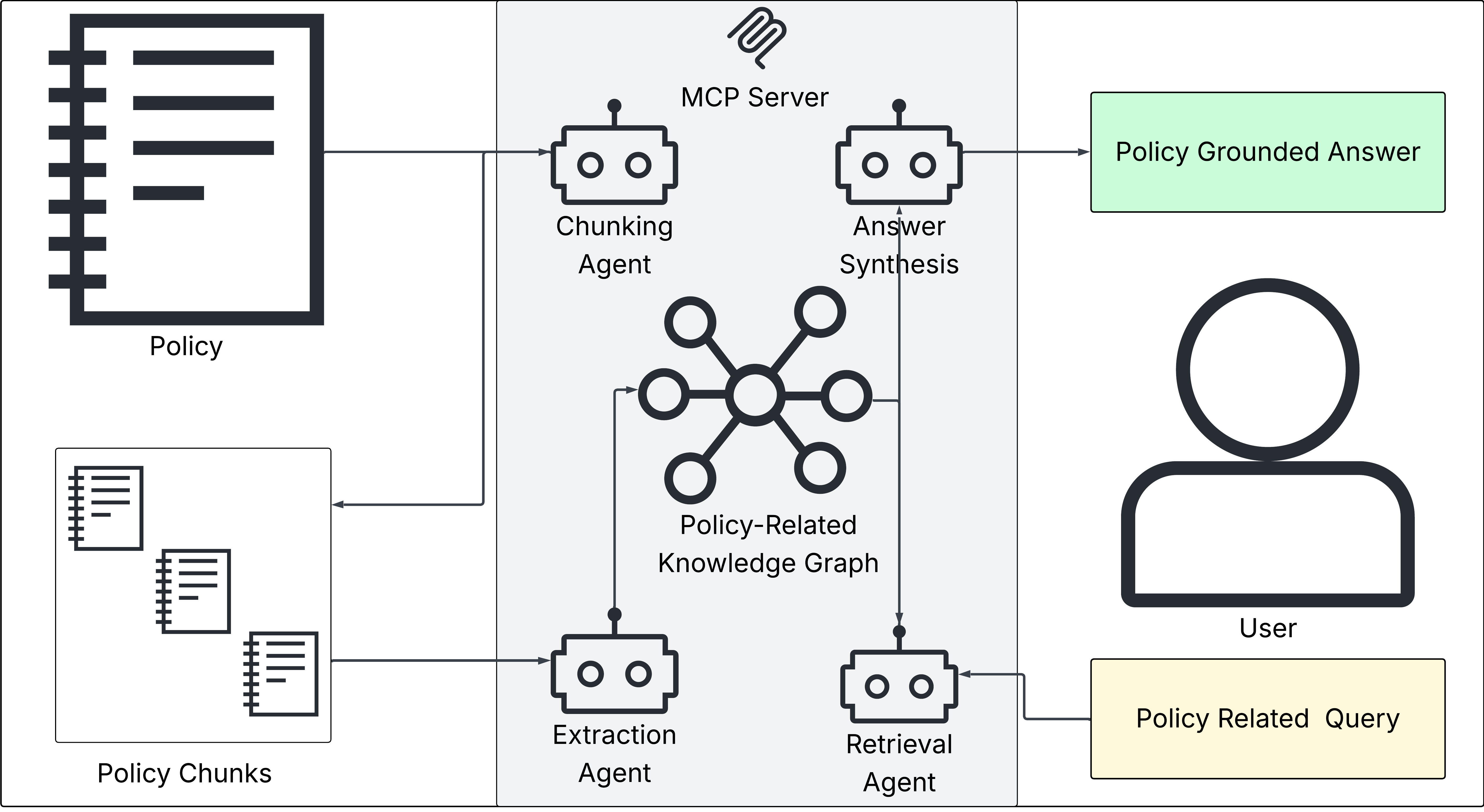}
\caption{Agentic framework overview. Policy documents are segmented by chunking agent, structured into KG by extraction agent, then traversed by a retrieval agent to synthesize policy-grounded answers.}
\label{fig:agentic_framework}
\end{figure*}

We develop a four-stage pipeline—chunking, extraction, retrieval, and answer synthesis—to investigate whether grounding LLM answers in a policy-derived KG improves accuracy in question-answering tasks, and if a formal ontology outperforms an emergent one. Figure \ref{fig:agentic_framework} shows an overview of our proposed framework. %for using LLMs to create policy-based KGs, which can then ground LLM reasoning in these policy texts. 
Our framework is exposed as a Model-Context-Protocol (MCP) server providing 16 tools, 7 resources, and 8 prompts, enabling integration with any MCP-compatible client. We provide a complete replication package with the MCP server, all prompts and evaluation question, as well as the results we report on in our anonymous GitHub repository. \footnote{\label{fn:repo}\scriptsize\url{https://anonymous.4open.science/r/LLM-Policy-Graph-MCP}}

\subsection{Document Chunking}
\label{sec:chunking}
Policy documents must be segmented into chunks small enough for the LLM's context window while preserving logical structure. Past work shows that structure-aware segmentation—splitting along logical boundaries rather than at fixed token counts—significantly improves retrieval accuracy and downstream QA performance  \cite{wang-etal-2025-document, akarajaradwong-etal-2025-nitibench, salim2025comparative}. Following these findings, we implement a \textbf{chunking agent} that identifies natural split boundaries using a two-phase \textit{scan + review} design.

In the \textbf{scan} phase, a sliding window (6{,}000~characters, 400-character overlap) advances through the document. At each position, we prompt the LLM with the current window, its absolute character offsets, and the location of the last boundary. The model returns candidate split points with short justifications (e.g., ``end of Article~10''), which we snap to the nearest paragraph or sentence break. In the \textbf{review} phase, we assemble the resulting boundary plan and check chunk sizes; if any chunk exceeds 4{,}000~characters, an additional LLM call proposes new splits for the oversized segments.

\subsection{Knowledge Graph Extraction}
\label{sec:extraction}

We process each chunk with an \textbf{extraction agent} to extract entities and relations into a typed KG. We evaluate two design choices: the choice of ontology schema and the strategy for incremental, multi-source extraction.

\subsubsection{Ontology Design}
\label{sec:ontology}

We compare two ontology schemas to investigate whether constraining the LLM to a predefined vocabulary improves extraction and downstream answering quality.

\paragraph{AIRO (ontology-derived closed).}
% We derive seven entity types and six relation types from the AI Risk Ontology (AIRO) \citep{golpayegani2022airo}, an OWL~2 ontology grounded in the EU AI Act and ISO risk management standards. Each type maps to one or more named OWL classes or object properties (e.g., \textsc{Risk\_Control} maps into \texttt{airo:RiskControl}, \textsc{Mitigates} into \texttt{airo:mitigatesRiskConcept}), with closely related subclasses collapsed into a single extraction label to reduce sparsity.

We ground our schema in the AI Risk Ontology (AIRO) \citep{golpayegani2022airo}, an OWL~2 ontology for AI risks built on the proposed EU AI Act (2021 draft) and ISO~31000/23894. We select seven OWL classes as entity types, collapsing subclasses into their nearest selected ancestor to reduce sparsity (e.g., \texttt{airo:Misuse}~$\to$~\textsc{Risk}; \texttt{airo:Hazard} and \texttt{airo:Threat}, which are subclasses of \texttt{airo:RiskSource}, are also mapped to \textsc{Risk}; \texttt{airo:Impact}~$\to$~\textsc{Consequence}). We then select six of AIRO's ${\sim}40$ object properties as relation types, choosing those that connect our seven entity types (e.g., \texttt{airo:mitigatesRiskConcept}~$\to$~\textsc{Mitigates}, \texttt{airo:hasConsequence}~$\to$~\textsc{Has\_Consequence}), and encode their OWL domain--range constraints as source$\to$target direction rules in the extraction prompt.

\paragraph{Open (emergent).}
In contrast to the closed AIRO schema, we impose no predefined type vocabulary. Instead, we instruct the LLM to assign descriptive \texttt{snake\_case} labels (e.g., \texttt{technical\_risk}, \texttt{regulatory\_obligation}).  A schema validator normalizes all labels to \texttt{lowercase\_with\_underscores} before storage to prevent formatting inconsistencies. During incremental extraction, we present the LLM with the existing type vocabulary to reuse established labels.
% rather than inventing synonyms. 

\subsubsection{Extraction Process}
We feed each chunk to the extraction agent in two passes: the first extracts entities, and the second extracts relations given the entity IDs from pass~1. For incremental, multi-source extraction, we augment the prompt with two forms of graph context: (1)~existing entity IDs and descriptions from the same policy source, encouraging the model to reuse established nodes; and (2)~high-value entities from other sources, enabling cross-policy relations at extraction time. We further deduplicate entities post-hoc using name-and-type matching.

All extraction prompts include: (1)~ICL examples from real EU AI Act text (Articles~10 and~14), (2)~relation direction constraints specifying source$\to$target types, and (3)~rules for splitting enumerated items into separate entities (e.g., each Annex~III category becomes its own node).    

% After extraction, we create cross-policy \textsc{Corresponds\_To} edges between entities from different sources. We identify candidate pairs by embedding cosine similarity ($\geq 0.70$, all-MiniLM-L6-v2), with a fallback to name-based string matching (SequenceMatcher~$\geq 0.80$), linking each entity to at most three cross-policy counterparts. Every entity records a \texttt{source\_chunk\_id} pointing to its originating chunk, and we store raw chunk texts alongside the graph so that original policy language can be retrieved during answer synthesis.

After extraction, we create cross-policy \textsc{Corresponds\_To} edges between entities from different sources. We identify candidate pairs by embedding cosine similarity ($\geq 0.70$, all-MiniLM-L6-v2), with a fallback to character-level name similarity ($\geq 0.80$) when embeddings are unavailable, linking each entity to at most three cross-policy counterparts. The cosine threshold of 0.70 follows prior work on semantic similarity for entity alignment \citep{reimers2019sentencebert}, while the string similarity threshold of 0.80 is consistent with near-duplicate detection baselines in entity resolution \citep{christophides2021entityresolution}. Every entity records a \texttt{source\_chunk\_id} pointing to its originating chunk, and we store raw chunk texts alongside the graph so that original policy language can be retrieved during answer synthesis.

\subsection{Retrieval and Answer Synthesis}
\label{sec:retrieval}

Given a question and a knowledge graph, the retrieval component produces a grounded answer through an adaptive two-path architecture.
The design draws on the ReAct framework \citep{yao2023react} and graph-grounded exploration \citep{sun2024thinkongraph}, with a complexity router that selects between a fast direct path and a full agent loop.

A \textbf{routing agent} classifies each question as \textit{direct} or \textit{agent} based on the question text and a summary of the graph schema. Simple questions (definition lookups, single-article queries, enumeration—task types T1--T3) are meant to take the \textbf{direct path}: we embed the question, retrieve the top-5 entities by cosine similarity, expand 1-hop neighbors from the top-3 seeds, and pass the assembled context to a single synthesis call (2~LLM calls total). This retrieve then expand design follows a two-stage pattern common in graph-augmented retrieval \citep{sun2024thinkongraph}: an initial recall set ($k$=5 balances coverage against context-window cost, consistent with the top-$k$ range of 3--10 used in dense retrieval benchmarks \citep{karpukhin2020dpr}), followed by 1-hop expansion from the highest-ranked seeds ($n$=3) to capture relational context without excessive fan-out. Complex questions (multi-hop reasoning, cross-policy linking, compliance checking—T4--T6) enter the \textbf{agent path}: a \textbf{traversal agent} explores the graph using native tool-calling in a ReAct loop (max 7~steps) with five graph tools—keyword search, semantic search, neighbor expansion, entity detail, and path finding—plus a terminal \texttt{synthesize\_answer} tool that collects evidence IDs and ends the loop. The 7-step budget is consistent with the bounded exploration budgets used in Think-on-Graph \citep{sun2024thinkongraph}.

Both paths converge on a \textbf{synthesis agent}---a single LLM call that receives five inputs: (1)~evidence entities in a relations-first serialization~\citep{dai2024kgunderstand}---relations grouped by type with source$\to$target labels, followed by entity details (ID, name, article reference, description, policy quote); (2)~raw chunk text from the source chunks that produced the evidence entities; (3)~two ICL examples selected by embedding similarity from a pool of 17 spanning all six task types; (4)~the original question; and (5)~task-specific instructions (e.g., cite article references, answer Yes/No first for compliance questions).

\subsubsection{Experimental Conditions}

We evaluate three conditions: \textbf{No-context baseline (NC):} the synthesis agent receives only the question and ICL examples---no KG context or chunk text---measuring base LLM knowledge alone. \textbf{AIRO KG:} the full pipeline (routing $\to$ direct/agent retrieval $\to$ chunk injection $\to$ synthesis) using the AIRO ontology graph. \textbf{Open KG:} the same pipeline using the open (emergent) ontology graph. Each condition is evaluated over five independent runs to assess variance.

\subsection{Evaluation}
\label{sec:evaluation}

We define six task types with increasing reasoning complexity, grounded in two complementary research traditions: KG comprehension benchmarks \citep{markowitz2025kgllmbench} and legal NLP task taxonomies \citep{guha2023legalbench}.

\paragraph{Task selection rationale.}
T1--T3 are inspired by the KG comprehension tasks in KG-LLM-Bench \citep{markowitz2025kgllmbench}, which shows that neighbor retrieval (our T2) and entity attribute QA (our T3) exhibit the largest variance across serialization formats---making them the most informative for our format comparison.
T4 extends multi-hop reasoning benchmarks \citep{yang2018hotpotqa, trivedi2022musique} to the policy domain, requiring chain reasoning across entities.
T5 (compliance checking) adapts binary classification tasks from LegalBench \citep{guha2023legalbench} to our setting: given a code snippet and a policy clause, the model must judge compliance (Yes/No/Partially), testing whether the KG provides sufficient regulatory grounding for applied decisions.
T6 (cross-policy reasoning) is proposed to challenge the framework with growing policies, as may be used in a live environment.

\subsubsection{Dataset Construction}

We construct 42 evaluation questions across T1--T6 (8~T1, 8~T2, 7~T3, 6~T4, 8~T5, 5~T6), spanning three policy sources (EU AI Act, NIST AI RMF, OWASP LLM Top~10).
T1--T4 questions are generated semi-automatically from KG queries (entity lookups, relation traversals, path queries) and verified by the authors against the source policy text.
T5 questions pair code snippets with binary compliance judgments, grounded in specific policy clauses.
T6 questions require mapping concepts across independently authored frameworks (e.g., ``Which OWASP LLM risks correspond to EU AI Act transparency requirements?'').

% Ground truth answers are derived directly from the knowledge graph for T1--T4 (ensuring a KG-lookup baseline scores 100\%) and from expert annotation for T5--T6.
% This design follows the evaluation methodology of KG-LLM-Bench \citep{markowitz2025kgllmbench}, where ground truth is grounded in the graph structure, ensuring that observed performance differences reflect the model's ability to reason over the provided representation rather than its parametric knowledge.

All reference answers are written and reviewed by the authors directly from the source policy documents rather than from any model-generated KG, ensuring that the ground truth is independent of the extraction pipeline under evaluation.
For T1--T4, we identify the relevant policy passage (e.g., Article~6 for high-risk AI system definitions), write the reference answer from that passage, and record the verbatim excerpt alongside each question so correctness can be independently verified.
For T5, the authors read the cited policy clauses and independently judge each code snippet against them, recording the relevant articles for traceability.
For T6, we manually identify cross-policy correspondences by reading both source documents, and then we verify that the expected mappings are supported by the text of each framework.
By comparing the KG conditions against the no-context baseline, we isolate the contribution of the knowledge graph and ensure that observed differences reflect the quality of the extraction and retrieval pipeline rather than the model's pre-existing policy knowledge, similar to prior work in KG-LLM benchmarking~\citep{markowitz2025kgllmbench}.

\subsubsection{Automated Scoring}

% Each question specifies a verification method matched to its task type:
Each evaluation question follows one of the following verification methods to gather a 0--1 score. 
\textbf{Semantic similarity} (T1, T3, T4)—word-overlap ratio between expected and generated answers.
\textbf{Set-match F1} (T2, T4)—keyword overlap scoring; for each expected item, we check for the exact substring match, then fall back to matching at least 2 distinctive words ($>$4~characters).
\textbf{Binary match} (T5)—we strip prefixes and check the first word for ``yes'' or ``no'', falling back to signal-phrase detection.
\textbf{Mapping accuracy} (T6)—for each expected cross-policy mapping, at least 2 key terms appear in the response.

\subsubsection{LLM-as-Judge}

As a complementary evaluation, we use a separate judge model (gpt-4.1-mini, temperature~0.1 to reduce variability) to score each response independently of the retrieval pipeline.
The judge receives the authoritative policy excerpt, the verified reference answer, and the model response.
It scores three dimensions (1--5 each)—\textit{accuracy} (factual correctness against the reference and source text), \textit{completeness} (coverage of required points, with paraphrases counting), and \textit{relevance} (whether the response directly answers the question)—and we report the mean as the composite score.
We judge each condition independently for every question across five replicated runs.
\section{Results}
\label{sec:results}

We evaluate five LLMs of three capability tiers: frontier API models (gpt-5-mini, gpt-4.1-mini), mid-range open-weight models (nemotron:30b, gpt-oss:20b), and a small local model (granite4:micro, 3B). Each was evaluated across three conditions: no-context baseline (NC), AIRO ontology-grounded KG (AIRO), and open emergent-schema KG (Open).
Each model builds its own KGs and uses them for retrieval, evaluating the full end-to-end pipeline.
Each condition is evaluated on 42 questions spanning six task types (T1--T6) over five runs. We report the mean $\pm$ std.

\subsection{RQ1: KG-Augmented Policy Reasoning}

\paragraph{Overall performance.}
Table~\ref{tab:overall} shows that KG augmentation improves judge scores for all five models.
The strongest heuristic gains appear with gpt-5-mini (+.120 AIRO, +.128 Open), while the largest judge improvement is for gpt-oss:20b (+0.46 AIRO, +0.55 Open).
% The judge scores are more lenient than the heuristic scorer---rewarding paraphrased correct answers that word-overlap metrics penalize---but the relative ordering of conditions is broadly consistent.
% Nemotron:30b shows a heuristic regression with KG augmentation ($-$.027 AIRO, $-$.050 Open) despite judge improvement score (+0.20 AIRO, +0.17 Open).
The judge consistently scores higher than the heuristic because it recognizes correct paraphrases that word-overlap metrics miss. For example, nemotron:30b scores $-$.027 heuristic on AIRO (apparent regression) but +0.20 on judge: its KG-grounded answers are factually correct but use different vocabulary than the reference, penalizing word overlap while satisfying the judge's accuracy criterion. Despite this calibration gap, both metrics agree on the key finding---KG $>$ NC---for all five models, and agree on condition ranking (which schema is better) for four of five.
Nemotron:30b shows a heuristic regression with KG augmentation ($-$.027 AIRO, $-$.050 Open) despite judge improvement score (+0.20 AIRO, +0.17 Open).

\begin{table*}[t]
\centering
\scriptsize
\begin{tabular}{|l|l|cc|cc|cc|cc|cc|cc|cc|}
\hline
\textbf{Model} & \textbf{Cond.} & \multicolumn{2}{c|}{\textbf{T1}} & \multicolumn{2}{c|}{\textbf{T2}} & \multicolumn{2}{c|}{\textbf{T3}} & \multicolumn{2}{c|}{\textbf{T4}} & \multicolumn{2}{c|}{\textbf{T5}} & \multicolumn{2}{c|}{\textbf{T6}} & \multicolumn{2}{c|}{\textbf{Overall}} \\
 & & \textbf{H} & \textbf{J} & \textbf{H} & \textbf{J} & \textbf{H} & \textbf{J} & \textbf{H} & \textbf{J} & \textbf{H} & \textbf{J} & \textbf{H} & \textbf{J} & \textbf{H} & \textbf{J} \\
\hline
\multirow{3}{*}{\textbf{gpt-5-mini}}
 & NC   & .59 & 4.3 & \textbf{.77} & 4.4 & .76 & 4.0 & .61 & 4.7 & .93 & 5.0 & .86 & 3.7 & .75\tiny{±.01} & 4.38\tiny{±.03} \\
 & AIRO & \textbf{.91} & 4.6 & .80 & 4.4 & .86 & 4.3 & .70 & 4.7 & \textbf{1.00} & 5.0 & .92 & 4.2 & .87\tiny{±.01} & 4.56\tiny{±.04} \\
 & Open & .89 & 4.9 & .77 & 4.2 & \textbf{.93} & 4.6 & \textbf{.73} & 4.8 & \textbf{1.00} & 5.0 & \textbf{.93} & 4.6 & \textbf{.88}\tiny{±.02} & \textbf{4.68}\tiny{±.02} \\
\hline
\multirow{3}{*}{\textbf{gpt-4.1-mini}}
 & NC   & .53 & 3.9 & \textbf{.68} & 3.9 & .64 & 3.6 & .53 & 4.0 & .90 & 4.6 & .88 & 3.0 & .69\tiny{±.01} & 3.90\tiny{±.05} \\
 & AIRO & \textbf{.92} & 4.7 & .64 & 4.0 & .87 & 4.1 & .56 & 3.7 & \textbf{1.00} & 4.9 & \textbf{.91} & 3.3 & .82\tiny{±.00} & 4.19\tiny{±.05} \\
 & Open & .91 & 4.7 & .63 & 3.9 & \textbf{.88} & 4.1 & \textbf{.63} & 4.2 & \textbf{1.00} & 4.8 & .90 & 3.3 & \textbf{.83}\tiny{±.00} & \textbf{4.23}\tiny{±.04} \\
\hline
\multirow{3}{*}{\textbf{gpt-oss:20b}}
 & NC   & .47 & 3.5 & \textbf{.64} & 3.4 & .56 & 3.0 & .52 & 3.6 & \textbf{1.00} & 4.8 & .81 & 2.8 & .66\tiny{±.02} & 3.79\tiny{±.07} \\
 & AIRO & .87 & 4.5 & .61 & 3.8 & .78 & 3.9 & .58 & 4.2 & .96 & 4.9 & .84 & 3.2 & .78\tiny{±.03} & 4.25\tiny{±.04} \\
 & Open & \textbf{.90} & 4.7 & .56 & 3.2 & \textbf{.89} & 4.5 & \textbf{.61} & 4.4 & .91 & 4.9 & \textbf{.87} & 3.4 & \textbf{.79}\tiny{±.01} & \textbf{4.34}\tiny{±.07} \\
\hline
\multirow{3}{*}{\textbf{nemotron:30b}}
 & NC   & .71 & 4.4 & \textbf{.85} & 4.2 & .79 & 3.9 & \textbf{.70} & 4.4 & \textbf{.93} & 4.9 & .92 & 3.7 & \textbf{.81}\tiny{±.02} & 4.29\tiny{±.04} \\
 & AIRO & \textbf{.91} & 4.9 & .73 & 4.1 & \textbf{.87} & 4.4 & .61 & 4.4 & .71 & 4.9 & .90 & 3.9 & .79\tiny{±.04} & \textbf{4.49}\tiny{±.07} \\
 & Open & .83 & 4.7 & .79 & 4.3 & .85 & 4.0 & .59 & 4.3 & .64 & 4.9 & \textbf{.91} & 4.0 & .76\tiny{±.05} & 4.45\tiny{±.10} \\
\hline
\multirow{3}{*}{\textbf{granite4:micro}}
 & NC   & .52 & 3.4 & .48 & 2.9 & .55 & 2.6 & \textbf{.51} & 3.5 & .62 & 4.2 & \textbf{.87} & 2.7 & .58\tiny{±.01} & 3.29\tiny{±.05} \\
 & AIRO & \textbf{.73} & 3.9 & .38 & 2.7 & \textbf{.72} & 3.6 & .50 & 3.4 & \textbf{.89} & 4.7 & \textbf{.87} & 2.8 & \textbf{.68}\tiny{±.01} & \textbf{3.56}\tiny{±.05} \\
 & Open & .72 & 3.9 & \textbf{.58} & 3.4 & .66 & 3.1 & .47 & 3.2 & .80 & 4.5 & .71 & 2.3 & .66\tiny{±.02} & 3.48\tiny{±.11} \\
\hline
\end{tabular}
\caption{Scores by model, condition, and task type (mean over 5 runs). H = heuristic (0--1), J = LLM-as-judge (1--5). Overall shows mean $\pm$ std. Each model uses its own KGs. Best H per model-task \textbf{bolded}.}
\label{tab:overall}
\end{table*}

\paragraph{Performance by task type.}
Table~\ref{tab:overall} presents per-task heuristic (H) and judge (J) scores for each model and condition, averaged over five runs.
T1 (entity lookup) and T3 (attribute retrieval) show the strongest and most consistent gains across all models, with heuristic improvements of +.11 to +.40.
These tasks benefit directly from verbatim policy quotes stored in KG entities, which align closely with expected answers.

T4 (multi-hop reasoning) shows model-dependent gains: gpt-5-mini improves by +.12 (Open) while granite4:micro shows slight regression ($-$.04), indicating that multi-hop graph traversal requires sufficient model capability to be effective.
% T2 (relation enumeration) shows modest or mixed gains.
% KG entities often describe related concepts in prose rather than enumerating discrete items, leading to partial recall on list-based expected answers.
T2 (relation enumeration) shows modest or mixed gains.
T2 questions expect enumerated lists (e.g., ``What categories are listed in Annex~III?''), but KG entity descriptions capture these as narrative summaries rather than individual items---for instance, a single entity may describe multiple Annex~III categories within one description field instead of storing each as a separate node with a \textsc{Listed\_In} relation. This mismatch between the KG's narrative structure and the expected list format leads to partial recall.
T5 (compliance checking) shows strong gains for weaker models: granite4:micro improves by +.27 (AIRO), and gpt-5-mini achieves perfect scores. Nemotron:30b exhibits a heuristic anomaly on T5 ($-$.22 AIRO, $-$.29 Open) despite near-perfect judge scores (4.9), as KG context causes verbose answers that fail the binary matcher. T6 (cross-policy reasoning) shows consistent improvements across models on judge scores.

\subsection{RQ2: Optimizing KG Creation and Retrieval}

\paragraph{Ontology schema.}
The open schema matches or slightly exceeds AIRO on judge scores for three of five models.
On heuristic scores, AIRO leads for nemotron:30b and granite4:micro, while the open KG schema leads for the rest.
The open schema's flexibility yields richer entity descriptions that align better with expected answers, while AIRO's formal structure may offer advantages (e.g., interoperability) not captured by our evaluation.

\paragraph{Model capability threshold.}
Granite4:micro benefits from KG on lookup tasks (T1: +.21, T3: +.17, T5: +.27) but regresses on complex tasks (T4: $-$.04, T6: $-$.16 Open).
Agentic traversal requires a multi-step tool that needs precise contexts for small LLMs.

\section{Discussion}
\label{sec:discussion}

\subsection{RQ1: KG-Augmented Policy Reasoning}

KG augmentation improves judge scores for all five models, but the mechanism varies by task type.
The strongest gains appear on T1 and T3 (+.11 to +.40 heuristic), where KG entities provide verbatim policy quotes that LLMs cannot reliably reproduce from parametric knowledge.
For example, on ``What is a high-risk AI system?'', the no-context gpt-5-mini vaguely cites ``Annex~II and III'' while the AIRO-grounded answer quotes Article~6(3) directly and enumerates all twelve provider obligations from Article~16(1)(a)--(l)---detail available only because the KG linked the ``High-Risk AI System'' entity to specific regulation and risk-control nodes.

T5 compliance questions show the starkest qualitative shift: without KG context, gpt-5-mini hedges (``Partially compliant''), lacking the specific article text; with AIRO context, it answers ``Yes'' definitively, citing Article~50(1) and Article~13 with verbatim quotes.
This pattern---from hedging to grounded certainty---appears across models, suggesting that LLMs possess regulatory reasoning ability but lack the specific textual evidence to commit to a judgment; the KG's role is therefore evidential, supplying the authoritative quotes that aid assessments with citable conclusions.

T6 cross-policy questions show the smallest benefit (+.01 to +.06).
On ``What requirements from all three policies apply to unbiased training data?'', the AIRO answer covers only the EU AI Act and NIST, missing the relevant OWASP LLM04 entity because ``poisoning'' and ``bias'' share little vocabulary.
The open-schema answer (+.11 over NC) succeeded via more diverse search paths, suggesting that cross-policy retrieval depends on vocabulary overlap between policy sources.

\subsection{RQ2: Optimizing KG Creation and Retrieval}

\paragraph{KG construction quality.}
Graph size ranges from 261 AIRO entities (granite4:micro) to 623 (nemotron:30b), but size does not predict downstream accuracy: gpt-5-mini's moderate graph (567 entities) outperforms nemotron's larger one because nemotron's open schema fragments into 47 entity types with many singletons (e.g., \textit{decommissioning\_practice}), diluting retrieval precision.
Cross-policy \textsc{Corresponds\_To} links correlate with T6 performance: gpt-5-mini's open schema has 42 cross-policy links (highest) and the best T6 scores, while gpt-oss:20b has only 4--6 and the weakest cross-policy retrieval.

\paragraph{Ontology schema design.}
Despite these per-model differences, the two schemas produce consistent structural patterns.
AIRO's closed vocabulary concentrates entities in RISK\_CONTROL (47--57\%) and relations in MITIGATES, yielding compact, navigable graphs.
The open schema distributes concepts more evenly but risks type proliferation---nemotron's 47 types versus gpt-5-mini's 21 demonstrate that model capability determines schema quality.
Downstream evaluation scores are close: the open schema matches or exceeds AIRO on judge scores for three of five models, with margins of 0.004--0.043.
The open schema's advantage appears to stem from more descriptive entity names (e.g., ``Harmful Effects of AI Systems [Article~1(1)]'' vs.\ the generic AIRO type ``CONSEQUENCE'') that overlap better with expected answer vocabulary. On ``How does risk management connect to human oversight?'', Granite's NC answer produces two paragraphs reasoning about Articles~26(5) and 79(1), while the AIRO-grounded answer collapses to three bullet points that restate facts without synthesizing the connection (0.42 vs.\ 0.52). Smaller LLMs may treat KG context as a constraint rather than evidence, suggesting that smaller models should bypass both the routing and agent steps entirely, using only the direct retrieval path.

\paragraph{Retrieval Accuracy.}
The direct path consistently retrieves 9--15 entities per question via semantic search and 1-hop expansion.
The agent path's effectiveness depends on model capability: gpt-5-mini averages 5.9 evidence entities per agent query, while gpt-oss:20b and granite4:micro average only 0.5---these smaller models fail to use the graph tools effectively, so their agent path provides almost no evidence.
For gpt-5-mini, the agent's selectivity improves performance: it matches or exceeds direct-path scores on T4 and T6 (+.05 to +.13) by following high-confidence relation chains rather than returning broad semantic matches.
On the hardest cross-policy question (``What requirements from all three policies apply to unbiased training data?''), only gpt-5-mini and gpt-4.1-mini successfully retrieve evidence from all three policy sources in the open schema; gpt-oss:20b and granite4:micro retrieve no evidence at all, falling back entirely on parametric knowledge.
The bottleneck for cross-policy retrieval is vocabulary overlap between policy sources: when policies use different terminology for the same concept (e.g., ``data poisoning'' vs.\ ``training data bias''), neither schema consistently bridges the gap.

\section{Conclusion and Future Work}
\label{sec:conclusion}

We presented an agentic framework that constructs knowledge graphs from AI governance documents and uses adaptive retrieval to ground LLM reasoning in authoritative policy text.
KG augmentation improves answer quality for all five models---up to +.13 heuristic F1 and +0.55 on LLM-as-judge---while an open, LLM-discovered ontology matches or exceeds a formal standards-derived schema.
We identify a model-capability threshold below which agentic graph traversal degrades rather than helps, suggesting that the retrieval strategy should be conditioned on the model that will use it. The capability threshold we observe for smaller LLMs suggests that policy optimization training~\cite{shao2024deepseekmath} or distillation could make KG-grounded risk detection viable on consumer hardware.

In the future, our plan is to extend the pipeline to source code---where agents jointly traverse a repository's AI feature graph and the policy KG---to test whether the retrieval mechanisms generalize to development scenarios. We will also conduct developer surveys to validate practical utility beyond automated metrics.
Cross-policy retrieval remains bottlenecked by vocabulary mismatch; entity alignment methods beyond embedding similarity (e.g., schema matching or LLM-based linking) may increase the generation of cross-policy edges that improve multi-policy reasoning. We will explore these challenges in the future. %The capability threshold we observe for smaller LLMs suggests that policy optimization training~\cite{shao2024deepseekmath} or distillation could make KG-grounded risk detection viable on consumer hardware.

\section{Limitations}
\label{sec:limitations}

Our 42-question evaluation spans three English-language Western AI governance frameworks, limiting generalizability: the small question set means individual items can disproportionately affect per-task scores, and we do not evaluate non-English regulatory texts or jurisdictions with different regulatory structures.
The heuristic scorer uses word-overlap rather than semantic similarity and may under-score correct answers with different vocabulary---as shown by nemotron:30b on T5, where the judge scores 4.9/5.0 but the binary heuristic penalizes verbose answers that bury the Yes/No signal.
Each model builds its KGs once; we do not evaluate variance in KG quality across extraction runs, which may confound cross-model comparisons. Finally, our evaluation is entirely automated; we do not yet include developer user studies to assess whether KG-grounded answers are actionable in practice or how they integrate into real compliance workflows.

\section{Ethical Considerations}
\label{sec:ethical_considerations}

This work uses publicly available policy documents (EU AI Act, NIST AI RMF, OWASP Top 10 for LLMs) and does not involve human participants or personal data.

% Limitations section required PrivateNLP
% \input{sections/limitations}

% Bibliography (acl.sty already sets \bibliographystyle{acl_natbib})
\bibliography{references}

\end{document}